\documentclass{article}

\usepackage{PRIMEarxiv}
\usepackage[utf8]{inputenc} 
\usepackage[T1]{fontenc}    
\usepackage{hyperref}       
\usepackage{url}            
\usepackage{booktabs}       
\usepackage{amsfonts}       
\usepackage{nicefrac}       
\usepackage{microtype}      
\usepackage{lipsum}
\usepackage{fancyhdr}       
\usepackage{graphicx}       
\graphicspath{{media/}}     
\usepackage{svg}
\usepackage{authblk}
\usepackage[paper=portrait,pagesize]{typearea}
\usepackage{multirow} 
\usepackage{makecell}

\usepackage{geometry}
\usepackage{multirow}
\usepackage{booktabs}
\usepackage{makecell}
\usepackage{pdflscape}

\usepackage{svg}

\usepackage{caption}

\usepackage{float}
\usepackage{amsmath}

\usepackage{xcolor,pifont}
\newcommand*\colourcheck[1]{%
  \expandafter\newcommand\csname #1check\endcsname{\textcolor{#1}{\ding{52}}}%
}
\colourcheck{blue}
\colourcheck{green}
\colourcheck{red}
\colourcheck{orange}

\usepackage{array}
\usepackage{ragged2e}

\usepackage{titlesec}
\titlespacing{\section}{0pt}{\parskip}{-\parskip}
\titlespacing{\subsection}{0pt}{\parskip}{-\parskip}
\titlespacing{\subsubsection}{0pt}{\parskip}{-\parskip}

\usepackage{tcolorbox}

\usepackage{amsmath}
\usepackage{algorithm}
\usepackage{algpseudocode}
\usepackage{amsfonts}
\usepackage{amsmath}
\usepackage{graphicx}
\usepackage{hyperref}

\usepackage{graphicx}
\DeclareGraphicsExtensions{.png}
\pdfpagewidth=\paperwidth
\pdfpageheight=\paperheight

\usepackage{graphicx}
\usepackage{float}
\usepackage{caption}

\title{Large Language Models for Patient Comments Multi-Label Classification}
  
\author[1]{\textbf{Hajar Sakai}}
\author[1]{\textbf{Sarah S. Lam}}
\author[2]{\textbf{Mohammadsadegh Mikaeili}}
\author[2]{\textbf{Joshua Bosire}}
\author[2]{\textbf{Franziska Jovin}}
\affil[1]{School of Systems Science and Industrial Engineering, Binghamton University, Binghamton, NY, USA}
\affil[2]{Cooper University Health Care, Camden, NJ, USA}

\begin{document}
\maketitle

\begin{abstract}
Patient experience and care quality are crucial for a hospital’s sustainability and reputation. The analysis of patient feedback offers valuable insight into patient satisfaction and outcomes. However, the unstructured nature of these comments poses challenges for traditional machine learning methods following a supervised learning paradigm. This is due to the unavailability of labeled data and the nuances these texts encompass. This research explores leveraging Large Language Models (LLMs) in conducting Multi-label Text Classification (MLTC) of inpatient comments shared after a stay in the hospital. GPT-4 Turbo was leveraged to conduct the classification. However, given the sensitive nature of patients’ comments, a security layer is introduced before feeding the data to the LLM through a Protected Health Information (PHI) detection framework, which ensures patients’ de-identification. Additionally, using the prompt engineering framework, zero-shot learning, in-context learning, and chain-of-thought prompting were experimented with. Results demonstrate that GPT-4 Turbo, whether following a zero-shot or few-shot setting, outperforms traditional methods and Pre-trained Language Models (PLMs) and achieves the highest overall performance with an F1-score of 76.12\% ± 0.021 and a weighted F1-score of 73.61\% ± 0.006 followed closely by the few-shot learning results. Subsequently, the results’ association with other patient experience structured variables (e.g., rating) was conducted. The study enhances MLTC through the application of LLMs, offering healthcare practitioners an efficient method to gain deeper insights into patient feedback and deliver prompt, appropriate responses.
\end{abstract}


\section{Introduction}
Patient experience is critical in assessing the quality of care delivered in hospitals, especially for inpatients. In recent years, its importance has progressively increased within the healthcare industry. This comes down to several reasons, among which multiples can be mentioned. Related literature (Doyle et al., 2013; Boulding et al., 2011; Senot et al., 2016) has repeatedly highlighted that patient experience is often associated with good health outcomes in terms of effective clinical adherence, lower readmission rates, and better financial performance. Additionally, it has also shown that patient-centered care promoted by providing a good experience to patients leads to the improvement of patient safety and the reduction of any potential risk that can occur during their hospital stay (Meterko et al., 2010; Shenoy, 2021). The relevance of patient experience has taken on another dimension with the introduction of the Hospital Value-Based Purchasing (VBP) Program (CMS, 2018). This initiative, led by the Centers for Medicare \& Medicaid Services (CMS), aims to adjust the payments delivered to hospitals by shifting the attention from considering only the quantity of care provided to including its quality as well. As a result, these rewards are intended to serve as an incentive to improve the care provided while lowering the costs. The VBP program assigns each acute care hospital a score based on five measures: Mortality and Complications, Healthcare-associated Infections, Patient Safety, Efficiency and Cost Reduction, and Patient Experience. Therefore, one way to achieve a higher score is to improve the patient experience. What better way to do that than listening to these same patients’ voices? 
\newpage
Patient feedback, when analyzed, can reveal the strengths and weaknesses of the assessed care services (Greaves et al., 2013; Manary et al., 2013). These experience narratives can often be collected through comments and surveys. In the case of a hospital stay, the type of survey addressed to patients after their discharge is the Hospital Consumer Assessment of Healthcare Providers and Systems (HCAHPS). This survey results from a collaboration between CMS and the Agency for Healthcare Research and Quality (AHRQ), which aims to understand patients’ perceptions of their experience at the hospital. It includes many sets of multiple-choice questions alongside a free-text section. The questions address diverse aspects of the care provided, which include interactions with different medical staff members and the hospital’s cleanliness. At the same time, the free-text section serves as a platform for patients to express their experiences using their own words. This leads to collecting unstructured patient feedback, which captures details not necessarily covered by the survey’s questions. Upon examination and structurization, the patients' comments can guide hospitals in improving their patient care.

\begin{figure}[H]
    \centering
    \includegraphics[scale=0.25]{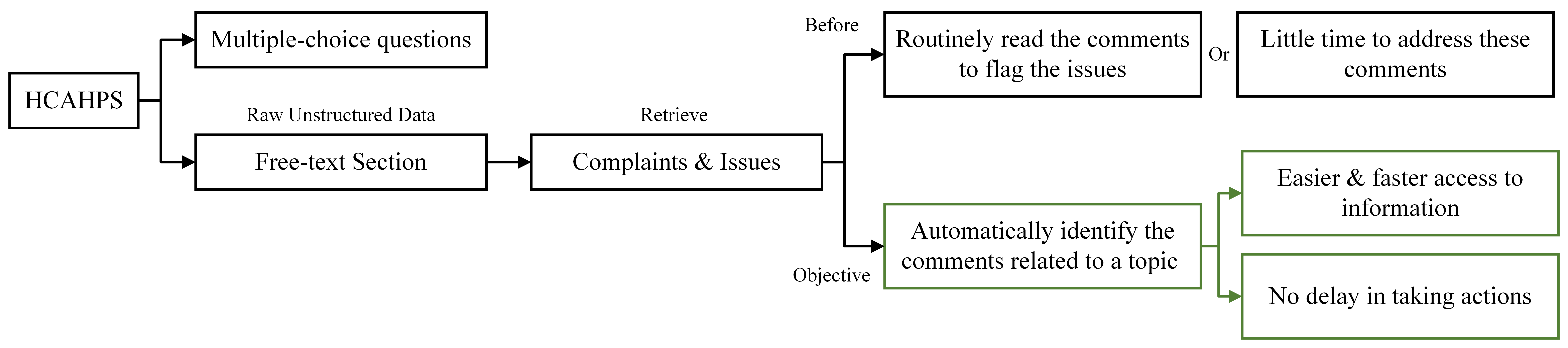}
    \caption*{\textbf{Figure 1.} Research Problem}
    \label{fig:research-prob}
\end{figure}

The healthcare industry is currently in the midst of a patient-centered era; therefore, understanding and acting on patient feedback has become key. These comments have an unstructured nature, which presents a significant challenge when dealt with. By combining linguistic and machine learning strengths, Natural Language Processing (NLP) provides an array of tools to enable patients' communication complexities decoding. MLTC emerges as a promising NLP technique for patient feedback analysis in this context, where multiple relevant topics are assigned to one comment. As a result, the multifaceted nature of patient experiences during their hospital stay will be reflected, especially in the highly influenced interactions with the hospital staff, the room amenities, the services provided, etc. However, implementing MLTC is challenging and requires developing efficient approaches to overcome its complexities. These challenges in a healthcare setting include dimensional complexity, where comments span many topics, making it necessary to develop a meticulous topic identification process customized to the studied hospital. This identification process would ideally be based on the typical patient’s journey mapping and collaboration with different hospital staff members. This would enable the identification of the different failure modes and thereby determine the relevant and actionable topics. A second challenge combines the class imbalance and the labeled data requirement to conduct an MLTC, for which an LLM-based approach is a potential solution. A third challenge is the concern of maintaining HIPAA (Health Insurance Portability and Accountability Act) compliance while classifying the comments using LLMs. Text mining techniques can be used for patients’ anonymization by developing a Protected Health Information (PHI) detection framework using text mining techniques.
\\[0.25cm]
The approaches leveraging LLMs vary depending on the resources available in terms of time, computing, and data. In healthcare, more specifically, an LLM can be developed from scratch: pre-trained and fine-tuned for the selected downstream tasks. Existing LLMs can also be fine-tuned on healthcare data. A third approach would rely on prompt engineering of existing general-purpose state-of-the-art LLMs. The latter can potentially be appropriate for healthcare MLTC since it is more cost and time efficient. There is minimal research on extracting insights from patient experience feedback. HCAHPS, for instance, has been the subject of a few research analyses (Gallan et al., 2022; Shovkun, 2018; Huppertz and Otto, 2018; Huppertz and Smith, 2014; Paul et al., 2022). However, none consists of conducting an MLTC where a pool of relevant topics is assigned (or not) to patients’ feedback. To the best of available knowledge, this is the first paper to carry out MLTC for HCAHPS comments using a LLM-based methodology. 
\\[0.25cm]
To tackle the challenges discussed above, this study makes the following key contributions: 
\begin{itemize}
    \item Topics Identification and Annotation Process: MLTC is not usually conducted for patient feedback mainly because no topics are predefined and can significantly vary from one hospital to another. As a result, no labels are available to train the model or evaluate the approach depending on the methodology used. This manuscript presents the topics’ identification process and the annotations done to evaluate the proposed approach. 
    \item PHI Detection Framework: Healthcare data is known for being sensitive and should be kept private. To de-identify the patients, a process that consists of text mining techniques (e.g., Regular Expressions) is developed to remove any PHI according to HIPAA. Additionally, all the LLMs involved are either used through their corresponding checkpoints or via API requests where data is not used for training, such as in the case of OpenAI’s GPT-4 Turbo (OpenAI, 2024a).
    \item LLM-based Approach: An unsupervised learning approach is used to overcome the data labeling, training, and potential class imbalance usually required for MLTC.
    \item For pattern identification, the MLTC outputs and their association with the patient’s demographics, information about their hospital stay, answers to the multiple-choice questions, and ratings are analyzed. This way, data triangulation is enabled by the cross-validation of the MLTC findings with the structured data collected, which enhances the reliability and credibility of the classification’s outputs.  
\end{itemize}

Based on the identified topics, the annotation of patient experience comments provides an evaluation set that involves five different annotators. Given the probabilistic nature of LLMs, to ensure the reliability of the classification, the evaluation and validation were done using three different samples of the dataset (100, 588, and 1,089 comments) and, thereby, different sample sizes and topic distributions. 
\\[0.25cm]
The paper is organized into the following sections: The “Related Work” section covers the existing literature focusing on the methods used to classify patients’ comments. In “Data and Annotation”, the topics identification process, annotation, and agreement analysis are presented. “Protected Health Information (PHI) Detection Framework” and “Methodology” encompass two parts of the MLTC methodology, which start with the PHI detection to ensure the patients’ de-identification to the LLM-based approach that enables the MLTC. The “Results and Discussion” section analyzes the results. Finally, “Conclusion and Future Directions” summarizes the research findings and suggestions for future research directions.

\section{Related Work}
Given the pivotal role of data-driven decisions in improving healthcare systems, the classification of patient comments using machine learning techniques emerges as a critical area of research. A literature review reveals that few research studies on the classification of patient experience feedback utilize HCAHPS as a data source, and even fewer apply machine learning techniques. This scarcity is notable, especially because such texts require prior manual annotation. Additionally, the private nature of healthcare textual data limits experimentation with different methods, particularly those involving LLMs. In this section, we review machine learning-based methodologies developed to classify patient comments in chronological order, regardless of the specific type of classification task. This approach allows us to trace the development of ideas and methodologies over time, which provides a clear picture of how the research has evolved in this area.
\\[0.25cm]
Because there is a lack of labeled patient comments to train a model, Mera and Ichimura (2008) tackles this issue by conducting diabetes patients’ comments MLTC using an unsupervised learning algorithm, Self-Organizing Feature Map (SOM). The MLTC concerns the anxiety type (Mental Problem, Physical Problem, Diet, Physical Activity, and Medicine) and the inner emotion (Happy, Sad, Fear, and Anger). Starting with the anxiety type classification, the text was first morphologically analyzed based on which ‘feature words’ strongly associated with each anxiety category were identified and quantified. Subsequently, these ‘feature words’ were considered while vectorizing the input text fed to the SOM. The training, which consists of adjusting the SOM node's weights, created a map where similar anxiety categories were clustered. Therefore, the mapping enabled the assignment of categories to patients’ comments. Similarly, the inner emotion classification was conducted. This research is among the earliest to explore the application of machine learning to the classification of patient comments, and its developed approach is more similar to topic modeling than typical text classification. A couple of years later, Alemi et al. (2012) explored conducting sentiment analysis followed by reasons for complaints classification of online patients’ comments expressed about pediatricians and obstetricians/gynecologists. Before the classification, manual labeling was conducted. Experiments were carried out using Decision Trees, Bagging, Support Vector Machine (SVM), and Multinomial Naïve Bayes. The complaints classification, particularly, led to the introduction of the time-to-next complaint measure. This metric could potentially enable tracking patient satisfaction over time. It would allow not only the quantification of patients’ feedback by monitoring the impact of any implemented improvements – an increase of the time-to-next might indicate positive feedback—but also provide a sense of the urgency of patient comments. Using online patient comments, Greaves et al. (2013) also conducted multiple sentiment analyses to interpret patient feedback. These binary classifications covered Overall Rating, Cleanliness, and Dignity and Respect. Experiments were conducted using supervised machine learning methods. However, instead of manual annotation, the patients’ ratings were converted to categorical variables and served as targets for training. Additionally, Wagland et al. (2016) analyzed free-text comments collected from colorectal cancer patient surveys through sentiment analysis. After manually labeling the comments, seven machine learning models (SVM, Random Forest (RF), DT, Generalized Linear Models Network (GLMNET), Bagging, Maxentropy, and Logitboost) were fitted and compared. This was followed by an analysis that contrasts the patient’s experience and their corresponding health-related quality of life (HRQoL) scores. As a result, an association between positive patient experiences and better HRQoL was found. Always intending to conduct sentiment analysis, Pan et al. (2018) addressed two key challenges when dealing with Chinese patient comments: word segmentation and polysemy. Character vector representations were generated instead of feeding word vectors to the Convolutional Neural Network (CNN) used for classification. The aim was to mitigate the errors that might result from word segmentation and polysemy by capturing more granular textual features. Additionally, the CNN employed was characterized by a convolution layer with multiple filters of varying sizes, which thus enables the extraction of a rich set of features from the comments. Furthermore, this paper introduced a segmentation pooling, where comments were divided into segments based on punctuation. Therefore, it allowed the model to capture the emotional nuances expressed in different parts of the comment while preserving the integrity of the comment’s sentiment. The changes in the CNN used led to a significant increase in accuracy compared to the traditional CNN. It is worth mentioning that, once again, manual labeling was resorted to. Often in the literature, sentiment analysis is coupled with topic modeling to retrieve useful information from patient comments. One example of these studies is Shah et al. (2018), in which the authors aimed to compare the patient comments about physicians shared in two different healthcare platforms. Besides using Online Multi-Objective Nonnegative Matrix Factorization (ONMF) for topic modeling, the authors compared multiple supervised classification models for the binary sentiment analysis: Naïve Bayes (NB), RF, DT (J.48), and Instance-Based K-Nearest Neighbors (Lazy.IBK). Later research (Hussain et al., 2020) explored the multiclass text classification (MCTC) of patient comments shared on a social media platform (Reddit). The target was to predict mental illness (Schizophrenia, Autism, Obsessive-Compulsive Disorder (OCD), and Post-Traumatic Stress Disorder (PTSD)). The comments were collected from each mental illness subreddit, and only clinical ones were considered. The labeling process was not explicitly explained; however, one can assume that the source subreddit was used for guidance. After preprocessing and vectorizing the text data using the Term Frequency-Inverse Document Frequency (TF-IDF), feature selection was conducted, and XGBoost was fitted. The performance of the chosen model was compared against NB and SVM, and it still outperformed them. Moving forward, Shan et al. (2021) explored a fine-grained sentiment analysis for patient comments: aspect-level sentiment analysis (ALSA). ALSA could extract the patient’s sentiment based on different aspects deemed relevant by healthcare professionals. This task was conducted using a hybrid approach that integrates Adversarial Sentiment Word Embeddings (ASWE), a pre-trained model (BERT), Attention Mechanism, and Bidirectional Long Short-Term Memory (BiLSTM). SEBERT, the proposed methodology, consisted of three main components. The first was the sentiment embeddings layer encompassing the words’ segmentation and removal of stopwords, followed by an ASWE. The output was afterward fed to a two-layer BiLSTM encoder. The results represented the sentiment information embedded in the patient’s comment. In parallel, the second component, the semantic embedding layer, was implemented using BERT. The third component, the output layer, integrated the sentiment and semantic features extracted by employing a multi-head self-attention mechanism, enabling the focus on the relevant information related to the sentiment and (predefined) aspect being analyzed. Manual labeling allowed the model’s evaluation, and as a result, its superiority over state-of-the-art methods was proven. While most of the papers discussed so far revolve around sentiment analysis, Asghari et al. (2022) investigated classifying comments collected from various social media to be categorized as either related to living organ donation or not. A web-scraping process was first developed, followed by a manual annotation. Subsequently, the collected text was preprocessed, and the classification was ensured by a Recurrent Neural Network (RNN). A year later, Sakai et al. (2023) explored the comments that can be collected from the HCAHPS, focus on classification of patient comments regarding food service issues using a neural network-based approach. The process consisted of manually labeling a sample of comments and leveraging BERT to conduct a flow augmentation on those labeled as those related to food service issues. Subsequently, the comments were preprocessed, and feature selection was combined with a sensitivity analysis to select the optimal vector size. Once the patients’ comments were adapted to a supervised learning approach, text classification was conducted using Linear SVM (Lin-SVM), Multilayer Perceptron (MLP), and Long Short-Term Memory (LSTM). Experimentation with different classification thresholds for the interpretability schedule in the case of both neural network-based models, LSTM outperformed the other models by achieving the least number of misclassifications. During the same year, Linton et al. (2023) leveraged Weakly Supervised Text Classification (WSTC) methods to analyze the colorectal cancer patients’ feedback collected from Patient-Reported Outcome Measures (PROMs) data. Five techniques were used and contrasted: BERTopic, CorEx Algorithm, Guided LDA (GLDA), WeSTClass, and X-Class. The aim was to classify the patients’ comments into predefined themes related to Health-Related Quality of Life (HRQoL), which results in an MLTC. In this proposed approach, the weak supervision was conducted by providing seed terms for the selected themes identified based on a literature review as input to the model and the patients’ comments. A manually annotated sample was used for the evaluation. Concurrently, another research (Alhazzani et al., 2023) emphasized using BERT-based approaches to conduct patients’ comments MLTC. After preprocessing the patients’ comments, feature extraction methods that result in static and dynamic word embeddings were explored. The objective consisted of the evolution from manually categorizing the comments into 25 predefined classes according to the Saudi Healthcare Complaints Taxonomy (SHCT) to leveraging deep learning-based models to enable an automated classification. This research used deep learning architectures that can be categorized into BiLSTM, Bidirectional Gated Recurrent Unit (BiGRU), or BERT-based models. In addition to fine-tuning each method, a BERT-based model, PX-BERT, was proposed. This model was tailored to patient experience data expressed in the Saudi dialect. Multiple evaluations included the consideration of all classes (25) and all sentiments and focused on negative comments while reducing the number of classes (20). This was done to investigate their effects on the classification’s performance. Later, in 2024, Sakai et al. (2024) employed advanced NLP techniques, combining BERT for comments embeddings, Neural Networks (i.e., MLP and LSTM) for binary sentiment analysis of HCAHPS comments, and Genetic Algorithm (GA) for hyperparameters optimization. The performance of the classification models was also compared against a baseline Linear-SVM. Results demonstrate that LSTM-GA slightly outperforms MLP-GA, while both significantly surpass the Linear-SVM baseline's performance. Always in 2024, patient comments’ sentiment classification that used openly available LLMs for Norwegian was explored and evaluated by Mæhlum et al. (2024). Similar to most of the previously discussed research papers, patients’ comments collection was followed by its manual annotation. The annotation was done on the comment and sentence levels, which led to two sentiment classifications, one binary that reflects the polarity (i.e., positive or negative) and the other categorical that reflects the intensity (i.e., slight, standard, or strong). Using two LLMs, ChatNorT5 and NorMistral, various prompt-based experiments were conducted. The experiments included both zero-shot and few-shot setups. The results were compared with human annotations and those of a Naïve Bayes, where a simple bag of words was used for text vectorization. However, one key challenge these LLMs encountered was handling neutral and mixed-polarity examples. This shows that the use of the current LLMs through prompt engineering could benefit from tailored enhancements. Involvement of LLMs is especially advantageous when using English text because most state-of-the-art LLMs have been developed and optimized primarily in English. Building on the groundwork laid by Mæhlum et al. (2024), this paper’s approach aims to refine the capabilities of LLMs by providing more context to the prompt and adjusting the outputs when necessary to not only better capture the corresponding sentiment (i.e., Positive Feedback vs other topics) but also identify the topics that patients are complaining about. Additionally, while the existing literature provides valuable insights, it is clear that the area of patients’ comments classification remains underexplored, particularly when integrating LLMs into the developed methodology.

\begin{landscape}
\begin{table}[h]
\captionsetup{labelformat=empty}
\centering
\caption*{\textbf{Table 1.} Machine Learning Applications for Patient Comments Classification}
\label{tab:classification_models}
\begin{tabular}{c c c c c c c c c c c}
 \toprule
\multirowcell{2}[0pt][c]{\centering\makecell{\textbf{Reference}}} & 
\multirowcell{2}[0pt][c]{\centering\makecell{\textbf{Language}}} & 
\multirowcell{2}[0pt][c]{\centering\makecell{\textbf{Data Source}}} & 
\multirowcell{2}[0pt][c]{\centering\makecell{\textbf{Classification} \\ \textbf{Type}}} & 
\multirowcell{2}[0pt][c]{\centering\makecell{\textbf{Best Model}}} & 
\multicolumn{6}{c}{\makecell{\textbf{Results of the Best}}} \\
\cmidrule(lr){6-11}
& & & & & \makecell{\textbf{Dataset/} \\ \textbf{Model/Label}} & \makecell{\textbf{Accuracy}} & \makecell{\textbf{F1} \\ \textbf{score}} & \makecell{\textbf{Precision}} & \makecell{\textbf{Recall}} & \makecell{\textbf{AUC} \\ \textbf{score}} \\

\midrule
\multirow{2}{*}{\makecell{\textbf{Mæhlum et al.} \\ (2024)}} & \multirow{2}{*}{Norwegian} & \multirow{2}{*}{NIPH Surveys} & \multirow{2}{*}{\makecell{Sentiment \\ Analysis \\ (Binary \& \\ Categorical)}} & \makecell{Binary: \\ CHATNORT5 \\ \& Zero-shot \\ (Prompt 3), \\ CHATNORT5 \\ \& Few-shot \\ (Prompt 2)} & Binary & - & 89.30\% & - & - & - \\
\cmidrule(lr){5-11}
 & & & & \makecell{Categorical: \\ CHATNORT5 \\ \& Zero-shot \\ (Prompt 3)} & Categorical & - & 42.40\% & - & - & - \\

\midrule
\makecell{\textbf{Sakai et al.} \\ (2024)} & English & HCAHPS & {\makecell{Binary \\ Classification}} & LSTM-GA & - & 91.50\% & 92.72\% & 91.71\% & 90.66\% & 94.90\% \\

\midrule
\multirow{6}{*}{\makecell{\textbf{Alhazzani et al.} \\ (2023)}} & \multirow{6}{*}{Arabic} & \multirow{6}{*}{\makecell{PX Surveys \\ 
(Negative \\ (13K), 25 \\ Classes/All- \\ Sentiments \\ (19K), 25 \\ Classes/All- \\ Sentiments \\(19K), 20 Classes)}} & \multirow{6}{*}{MLTC} & \multirow{6}{*}{\makecell{PX\_BERT, \\ AraBERTv02}} & \makecell{Negative (13K), \\ 25 Classes \\ PX\_BERT} & 55.14\% & 43.06\% & 63.61\% & 32.54\% & - \\
\cmidrule(lr){6-11}
 & & & & & \makecell{Negative (13K), \\ 25 Classes \\ AraBERTv02} & 60.24\% & 38.83\% & 81.22\% & 25.51\% & - \\
\cmidrule(lr){6-11}
 & & & & & \makecell{All-Sentiments \\ (19K), 25 Classes \\ PX\_BERT} & 55.84\% & 43.07\% & 67.45\% & 31.63\% & - \\
 \cmidrule(lr){6-11}
 & & & & & \makecell{All-Sentiments \\ (19K), 25 Classes \\ AraBERTv02} & 57.95\% & 47.10\% & 64.12\% & 37.22\% & - \\
 \cmidrule(lr){6-11}
 & & & & & \makecell{All-Sentiments \\ (19K), 20 Classes \\ PX\_BERT} & 55.74\% & 45.50\% & 55.64\% & 38.49\% & - \\
 \cmidrule(lr){6-11}
 & & & & & \makecell{All-Sentiments \\ (19K), 20 Classes \\ AraBERTv02} & 60.02\% & 48.70\% & 66.00\% & 38.59\% & - \\

\midrule
\makecell{\textbf{Linton et al.} \\ (2023)} & English & PROMs & MLTC & CorEx & - & 81.30\% & - & - & - & - \\

\bottomrule
\end{tabular}
\end{table}

\begin{table}[h]
\captionsetup{labelformat=empty}
\centering
\caption*{\textbf{Table 1.} Cont.}
\label{tab:classification_models}
\begin{tabular}{c c c c c c c c c c c}
 \toprule
\multirowcell{2}[0pt][c]{\centering\makecell{\textbf{Reference}}} & 
\multirowcell{2}[0pt][c]{\centering\makecell{\textbf{Language}}} & 
\multirowcell{2}[0pt][c]{\centering\makecell{\textbf{Data Source}}} & 
\multirowcell{2}[0pt][c]{\centering\makecell{\textbf{Classification} \\ \textbf{Type}}} & 
\multirowcell{2}[0pt][c]{\centering\makecell{\textbf{Best Model}}} & 
\multicolumn{6}{c}{\makecell{\textbf{Results of the Best}}} \\
\cmidrule(lr){6-11}
& & & & & \makecell{\textbf{Dataset/} \\ \textbf{Model/Label}} & \makecell{\textbf{Accuracy}} & \makecell{\textbf{F1} \\ \textbf{score}} & \makecell{\textbf{Precision}} & \makecell{\textbf{Recall}} & \makecell{\textbf{AUC} \\ \textbf{score}} \\

 \midrule
 \makecell{\textbf{Sakai et al.} \\ (2023)} & English & HCAHPS & \makecell{Binary \\ Classification} & LSTM & - & 95.51\% & 86.54\% & 72.69\% & 55.37\% & 58.21\% \\

\midrule
\multirow{4}{*}{\makecell{\textbf{Asghari et al.} \\ (2022)}} & \multirow{4}{*}{English} & \multirow{4}{*}{\makecell{NYT, \\ Reddit, \\ Twitter, \\ YouTube}} & \multirow{4}{*}{\makecell{Binary \\ Classification}} & \multirow{4}{*}{DNN} & \makecell{NYT} & 60.70\% & \makecell{70.00\% \\ (Macro)} & 51.10\% & 85.50\% & - \\
\cmidrule(lr){6-11}
 & & & &  & Reddit & 47.10\% & \makecell{58.00\% \\ (Macro)} & 34.20\% & 73.80\% & - \\
 \cmidrule(lr){6-11}
 & & & &  & Twitter & 46.80\% & \makecell{46.80\% \\ (Macro)} & 46.80\% & 15.00\% & - \\
  \cmidrule(lr){6-11}
 & & & &  & YouTube & 46.20\% & \makecell{61.20\% \\ (Macro)} & 61.20\% & 42.30\% & - \\

 \midrule
\multirow{2}{*}{\makecell{\textbf{Shan et al.} \\ (2021)}} & \multirow{2}{*}{Chinese} & \multirow{2}{*}{Haodaifu} & \multirow{2}{*}{ALSA} & \multirow{2}{*}{SEBERT} & Aspect & 93.00\% & - & - & - & - \\
\cmidrule(lr){6-11}
 & & & &  & Sentiment & 92.06\% & 92.73\% & 92.56\%  & 92.89\% & - \\

 \midrule
\multirow{4}{*}{\makecell{\textbf{Hussain et al.} \\ (2020)}} & \multirow{4}{*}{English} & \multirow{4}{*}{Reddit} & \multirow{4}{*}{MCTC} & \multirow{4}{*}{XGBoost} & OCD & \multirow{4}{*}{68.00\%} & 58.00\% & 87.00\% & 44.00\% & - \\
\cmidrule(lr){6-6} \cmidrule(lr){8-11}
 & & & &  & Autism & & 72.00\% & 69.00\% & 76.00\% & - \\
\cmidrule(lr){6-6} \cmidrule(lr){8-11}
 & & & &  & PTSD &  & 63.80\%  & 76.00\% & 54.00\% & - \\
\cmidrule(lr){6-6} \cmidrule(lr){8-11}
 & & & &  &  Schizophrenia &  & 70.0\%  & 60.00\% & 84.00\% & - \\

 \midrule
\multirow{2}{*}{\makecell{\textbf{Shah et al.} \\ (2018)}} & \multirow{2}{*}{English} & \multirow{2}{*}{\makecell{RateMDs \& \\ Healthgrades}} & \multirow{2}{*}{\makecell{Sentiment \\ Analysis \\ (Binary)}} & \makecell{NB, SVM} & RateMDs & \makecell{73.40\% \\ 93.60\%} & \makecell{ 65.00\% \\ 53.00\%} & \makecell{ 65.00\% \\ 53.00\%} & \makecell{65.00\% \\ 53.00\%} & \makecell{69.00 \% \\ 53.00 \%} \\
\cmidrule(lr){5-11}
 & & & & \makecell{RF, SVM} & Healthgrades & \makecell{90.90\%, \\ 75.00\%} & \makecell{54.00\%, \\ 62.00\%} & \makecell{61.00\%, \\ 63.00\%}  & \makecell{58.00\%, \\ 63.00\%} & \makecell{61.00\%, \\ 63.00\%}  \\

 \midrule
 \makecell{\textbf{Pan et al.} \\ (2018)} & Chinese & Haodaifu & \makecell{Sentiment \\ Analysis \\ (Binary)} & \makecell{Character \\ Vector Convolutional \\ Neural Network \\ with \\ Segmentation \\ Pooling}
 & - & 88.20\% & 88.20\% & 85.60\% & 86.90\% & - \\

 \midrule
 \makecell{\textbf{Wagland et al.} \\ (2015)} & English & PROMs & \makecell{Sentiment \\ Analysis \\ (Binary)} & SVM
 & - & - & 80.00\% & 83.50\% & 78.00\% & - \\

\bottomrule
\end{tabular}
\end{table}

\begin{table}[h]
\captionsetup{labelformat=empty}
\centering
\caption*{\textbf{Table 1.} Cont.}
\label{tab:classification_models}
\begin{tabular}{c c c c c c c c c c c}
 \toprule
\multirowcell{2}[0pt][c]{\centering\makecell{\textbf{Reference}}} & 
\multirowcell{2}[0pt][c]{\centering\makecell{\textbf{Language}}} & 
\multirowcell{2}[0pt][c]{\centering\makecell{\textbf{Data Source}}} & 
\multirowcell{2}[0pt][c]{\centering\makecell{\textbf{Classification} \\ \textbf{Type}}} & 
\multirowcell{2}[0pt][c]{\centering\makecell{\textbf{Best Model}}} & 
\multicolumn{6}{c}{\makecell{\textbf{Results of the Best}}} \\
\cmidrule(lr){6-11}
& & & & & \makecell{\textbf{Dataset/} \\ \textbf{Model/Label}} & \makecell{\textbf{Accuracy}} & \makecell{\textbf{F1} \\ \textbf{score}} & \makecell{\textbf{Precision}} & \makecell{\textbf{Recall}} & \makecell{\textbf{AUC} \\ \textbf{score}} \\

\midrule
\multirow{3}{*}{\makecell{\textbf{Greaves et al.} \\ (2013)}} & \multirow{3}{*}{English} & \multirow{3}{*}{\makecell{Online Patients' \\ Comments}} & \multirow{3}{*}{\makecell{Sentiment \\ Analysis \\ (Binary)}} & \multirow{3}{*}{MNB} & Overall Rating & 88.60\% & 89.00\%  & - & - & 94.00\% \\
\cmidrule(lr){6-11}
 & & & &  & Cleanliness & 81.20\% & 84.00\%  & - & - & 88.00\% \\
 \cmidrule(lr){6-11}
 & & & &  & \makecell{Dignity \\ \& Respect} & 83.70\% & 85.00\%  & - & - & 91.00\% \\

 \midrule
\multirow{9}{*}{\makecell{\textbf{Alemi et al.} \\ (2012)}} & \multirow{9}{*}{English} & \multirow{9}{*}{\makecell{Online Patients' \\ Comments}} & \multirow{9}{*}{MLTC} & Bagging & Praise & - & 100\%  & 89.00\% & 95.00\% & 74.00\% \\
\cmidrule(lr){5-11}
 & & & & Bagging   & Complaint & - & 53.00\%  & 70.00\% & 71.00\% & 83.00\% \\
 \cmidrule(lr){5-11}
 & & & &  SVM  & \makecell{Doctor gives \\ good advice \\ \& treatment} & - & 70.00\%  & 68.00\% & 72.00\% & 80.00\% \\
 \cmidrule(lr){5-11}
 & & & &  Bagging & \makecell{Doctor takes \\ enough time} & - & 67.00\%  & 84.00\% & 56.00\% & 89.00\% \\
 \cmidrule(lr){5-11}
 & & & &   DT, SVM & \makecell{Doctor explains \\ well} & - & \makecell{70.00\%, \\ 68.00\%}  & \makecell{70.00\%, \\ 70.00\%} & \makecell{70.00\%, \\ 66.00\%} & \makecell{81.00\%, \\ 90.00\%} \\
  \cmidrule(lr){5-11}
 & & & &  Bagging & Staff related & - & 87.00\%  & 91.00\% & 81.00\% & 94.00\% \\
  \cmidrule(lr){5-11}
 & & & &  Bagging  & \makecell{Staff friendly \\ and helpful} & - & 45.00\%  & 67.00\% & 35.00\% & 89.00\% \\
  \cmidrule(lr){5-11}
 & & & & DT &  Doctor listens & - & 74.00\%  & 77.00\% & 77.00\% & 88.00\% \\
  \cmidrule(lr){5-11}
 & & & & SVM & Wait related & - & 53.00\%  & 41.00\% & 73.00\% & 89.00\% \\

\midrule
    \multirow{6}{*}{\makecell{\textbf{Mera and Ichimura} \\ (2008)}} & \multirow{6}{*}{Japanese} & \multirow{6}{*}{\makecell{Health \\ Support \\ Intelligent \\ System for \\ Diabetic \\ Patients \\(HSISD)}} & \multirow{6}{*}{MLTC} & \multirow{6}{*}{SOM} & \makecell{Mental \\ Problem} & 87.50\% & - & - & - & - \\
\cmidrule(lr){6-11}
 & & & &  & \makecell{Physical \\ Problem} & 66.67\% & - & - & - & - \\
 \cmidrule(lr){6-11}
 & & & &  & Diet & 75.00\% & - & - & - & - \\
 \cmidrule(lr){6-11}
 & & & &  & \makecell{Physical \\ Activity} & 88.89\% & - & - & - & - \\
 \cmidrule(lr){6-11}
 & & & &  & Medicine & 80.00\% & - & - & - & - \\
 \cmidrule(lr){6-11}
 & & & &  & Inner Emotion & 63.16\% & - & - & - & - \\

\bottomrule
\end{tabular}
\end{table}
\end{landscape}

\section{Data and Methodology}
Patient feedback MLTC, even when conducted using LLMs, still needs to be validated. This requires samples’ manual annotations after label identification. Once completed, the patients should be de-identified from the comments. Subsequently, classification can be carried out and, after that, evaluated. The details of this proposed approach are detailed in this section and presented in Figure 2.

\begin{figure}[H]
    \centering
    \includegraphics[scale=0.7]{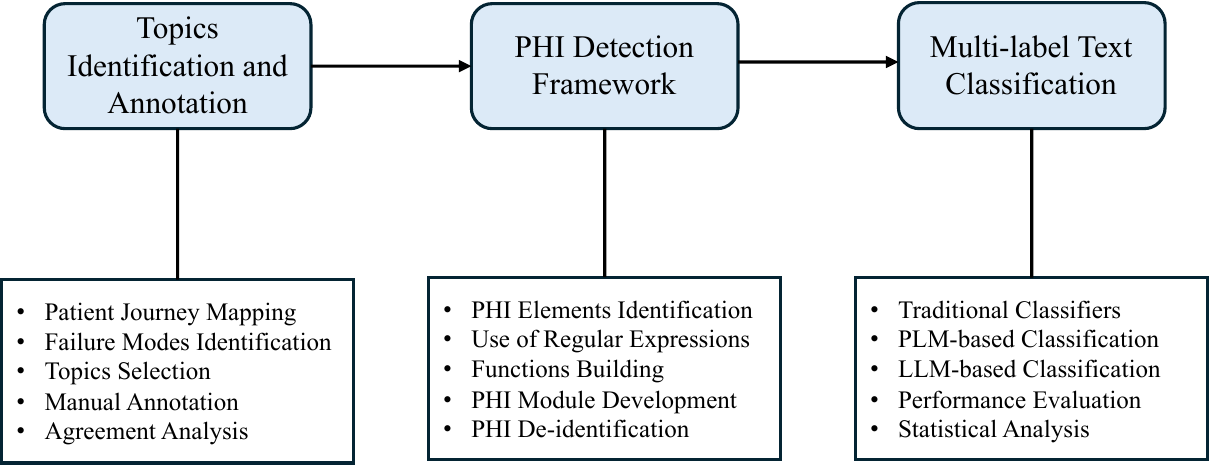}
    \caption*{\textbf{Figure 2.} Research Methodology}
    \label{fig:mltc-approaches}
\end{figure}

\subsection{Data and Annotation}
Assessing the effectiveness of the proposed methodology for patient feedback MLTC requires a robust evaluation process. This process can be based on comparing the LLM-based against existing techniques using annotated data. However, before annotating, it is essential first to define the relevant topics (i.e., labels) that will form the basis of the classification.

\subsubsection{Topics Identification}
Patients’ feedback can fall under the umbrella of various topics regarding MLTC. However, having a significantly large number of topics would lead to dealing with an Extreme Multi-Label Text Classification case. This task comes with various challenges, among which the sparsity of label occurrences (Babbar and Schölkopf, 2017) and evaluation complexity (Bhatia et al., 2015) can be mentioned. Additionally, the relevant topics vary from one hospital to another depending on the priority, importance, and feasibility of action of the topic.

\begin{figure}[H]
    \centering
    \includegraphics[scale=0.35]{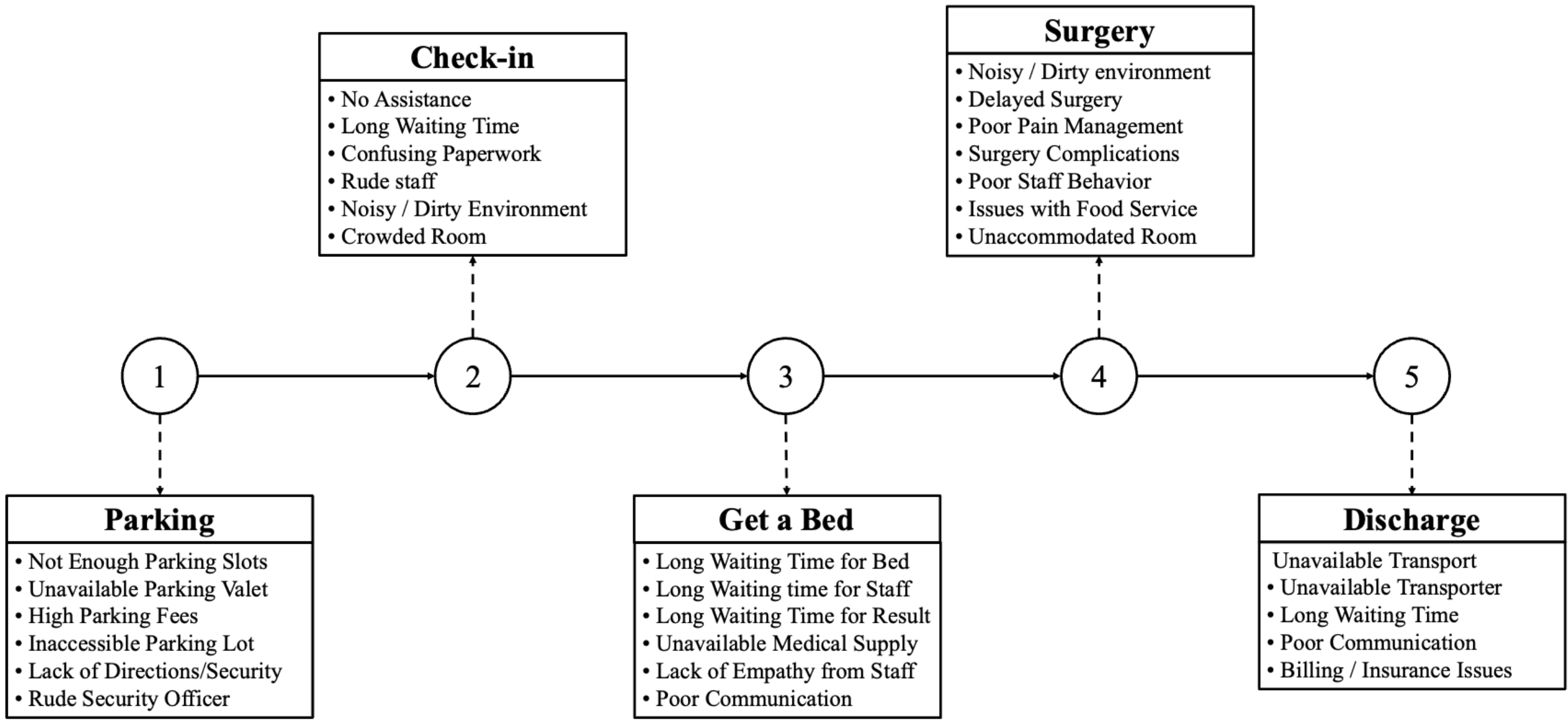}
    \caption*{\textbf{Figure 3.} Patient’s Journey Mapping in the Case of an Elective Surgery}
    \label{fig:mltc-approaches}
\end{figure}

Identification of classification topics starts with mapping a typical patient’s journey in the case of different potential scenarios that involve a hospital stay. Figure 3 illustrates the case of a patient who visits a hospital for elective surgery from the moment the hospital parking lot is reached until discharge. The patient can encounter multiple failure modes during each step of their journey. These failure modes give an understanding of the different potential areas of improvement and, therefore, the topics under which the comments should be classified. These failure modes were carefully examined and discussed. After multiple rounds of topic identification and discussion, ten topics for MLTC were selected: Positive Feedback, Noisy Environment, Missing Personal Belongings, Miscellaneous, Staff-related Issues, Long Waiting Time, Issues with Food Service, Room-related Issues, Medical-related Issues, Discharge-related Issues. Except for the first topic, all the remaining ones have a negative connotation. This decision targeted negative issues by analyzing patients' complaints, which thereby would help in the identification of areas that need improvement without an explicit focus on positive topics. 

\subsubsection{Annotation and Agreement Analysis}
In the case of this research, the annotated dataset is used to evaluate and validate the approach. The dataset consists of a set of patient comments randomly sampled from those received in the HCAHPS between 2019 and 2023 by a hospital located in New Jersey, USA. These comments were selected to represent a broad set of patient experiences regarding various aspects of healthcare services. No stratified sampling was involved because no labels were available for such process. The resulting dataset consists of 1,089 comments.  

\begin{figure}[H]
    \centering
    \includegraphics[scale=0.5]{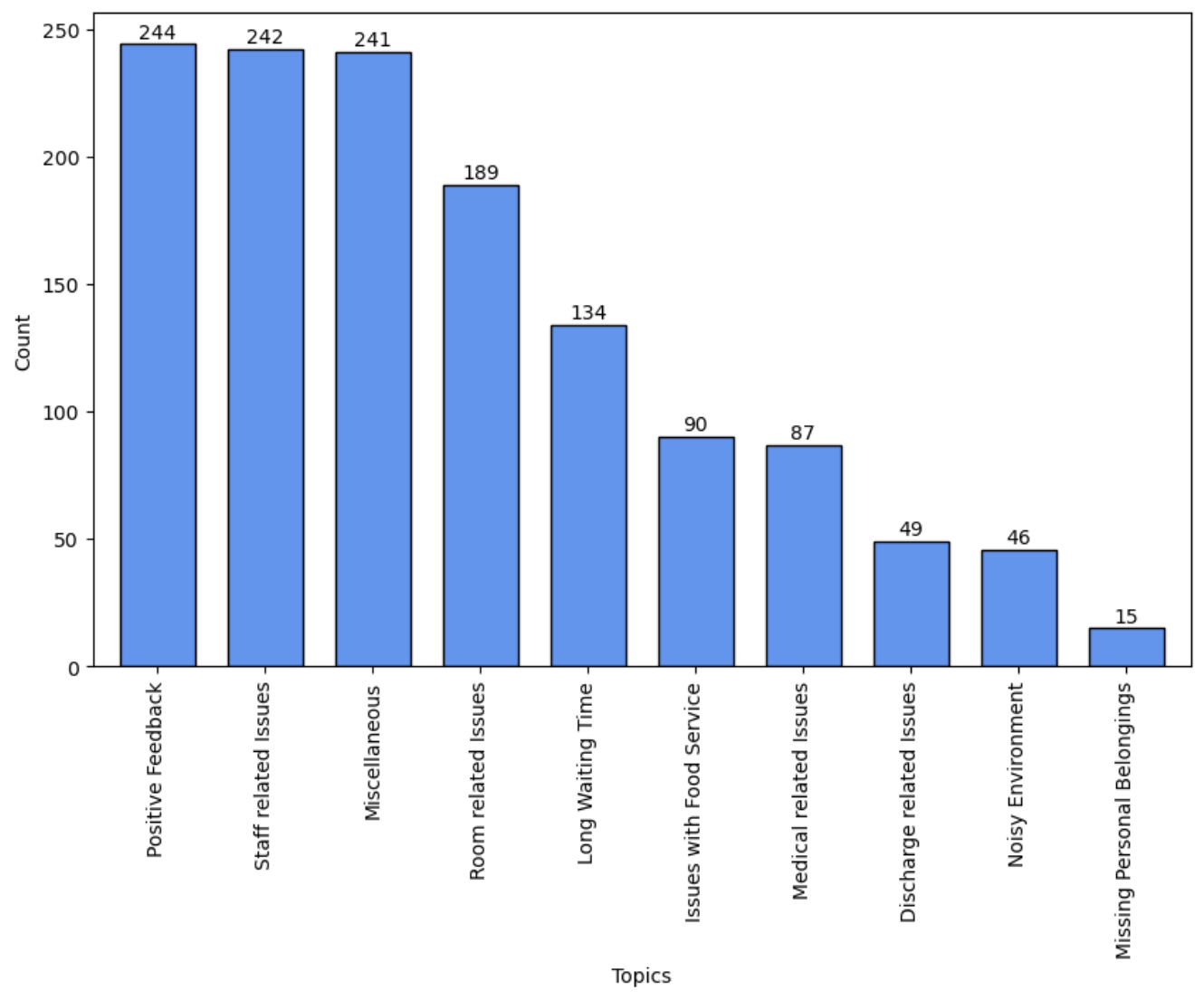}
    \caption*{\textbf{Figure 4.} Topics Distribution}
    \label{fig:mltc-approaches}
\end{figure}

The annotation process involved three rounds, in which each successive round incorporating additional comments. It was carried out by a team experienced in patient family-centered care, who categorized the comments into the predefined topics. Two annotators labeled each comment independently to enhance the reliability and quality of the annotated data. At the same time, the entire dataset’s annotation involved five annotators. Moreover, agreement analyses were carried out following each round to assess the consistency of the labeling conducted. Cohen’s Kappa statistic was used for each inter-annotator agreement analysis. This statistic was calculated for each comment to determine the degree of agreement across all labels. Comments with Kappa values less than 1 were subjected to a review discussion among the annotators to resolve discrepancies. These discussions served to both refine the topics’ definitions (Appendix 1) and make sure that all the annotators have the same understanding of the topics’ details and instructions. Figure 4 summarizes the distributions of the topics that result from the annotation. 

\subsection{Protected Health Information (PHI) Detection Framework}

With the digital transformation, healthcare data is growing exponentially, in which 80\% is unstructured (Kong, 2019). It has various sources (e.g., surveys, wearable devices) and comes in different formats (e.g., text, time series). In the case of this research, textual data is considered. Before inputting it to an LLM, PHI should be detected and masked, which, in other words, means that the comments should be manipulated. Text manipulation is, without a doubt, challenging, time-consuming, and exhausts resources. However, PHI detection tasks can be automated by developing functions based on the built-in methods contained in Python libraries such as RE (Regular Expression), NLTK (Natural Language Toolkit), Pyap, spaCy, Pandas, and NumPy. 
\\[0.25cm]
HIPAA considers a list of elements as PHI (HIPAA Journal, 2024) that we can summarize as follows: First and Last Names, Dates (e.g., Date of Birth (DOB)), Phone Numbers, Mailing Addresses, Email Addresses, Web URLs, and Numbers (e.g., Social Security Number (SSN)). Based on this, rules and regular expressions were developed inside each type’s function (or sub-function). Once a PHI is detected, it is redacted to conserve the comment's meaning while ensuring that the patient is de-identified.

\begin{figure}[H]
    \centering
    \includegraphics[scale=0.3]{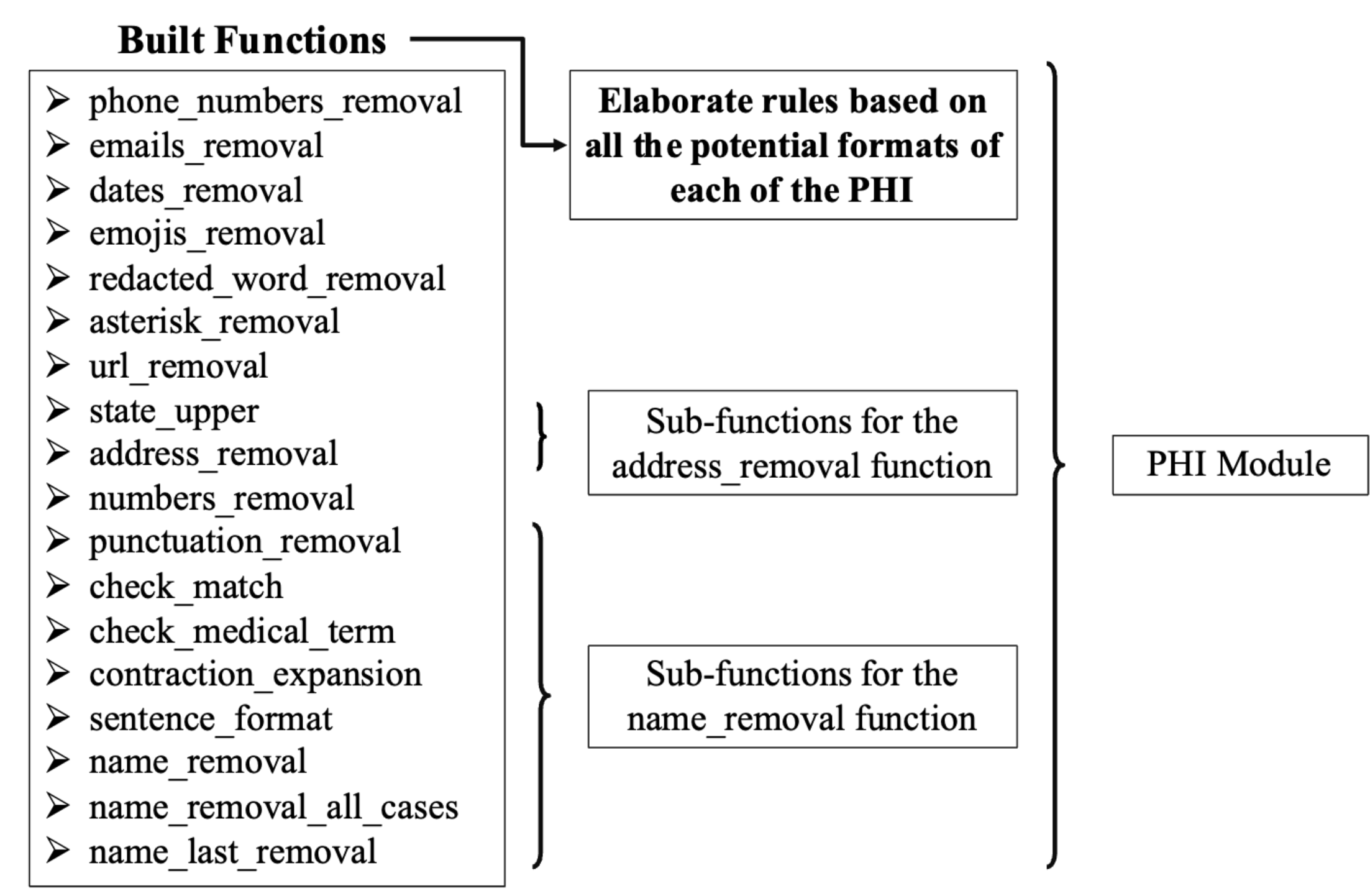}
    \caption*{\textbf{Figure 5.} PHI Detection Framework}
    \label{fig:mltc-approaches}
\end{figure}

Figure 5 summarizes the framework developed for PHI detection. Starting with the functions built to target each of the components of the comments that fall under HIPAA consideration for PHI, to creating a module that the end user can quickly call to detect and redact each PHI. Additionally, to completely anonymize the comments, the names of the doctors and the area’s hospitals were redacted. Subsequently, a sample of comments was selected for manual validation to ensure that the defined functions covered all possible scenarios. 

\subsection{LLM-based MLTC}
MLTC, in the context of this paper, is a task consisting of assigning relevant topics to a given comment shared by a patient in the HCAHPS. Unlike single-label classification, which can be binary or multi-class, MLTC permits one feedback to be associated with one or more labels from a predefined set. This research explores the efficiency of using a high-performing LLM, GPT-4 Turbo, for this purpose. This LLM was developed by OpenAI (2024b) and is a cost-optimized and faster version of GPT-4. In addition to an advanced natural language understanding and generation, this LLM is able to handle various NLP tasks such as text classification and summarization. It is also a general-purpose LLM, which means that it was trained on various text domains. Prompt engineering was employed to stay within the optic of conducting a compute-efficient MLTC. A prompt was carefully crafted to include the predefined set of topics and their corresponding definitions, in addition to the classification instructions (similar to those shared with the group of annotators). Experiments using 0-shot learning and few-shot learning (1-shot, 3-shot, and 5-shot) were performed. In parallel, classifications using traditional machine learning methods relying on vectorization and multi-label transformation techniques as well as PLMs (i.e., BERT and BART) were also made. Their respective results were evaluated and contrasted using multiple classification metrics. For individual topic evaluation, F1-score and AUC-score were employed. Macro, micro, and weighted F1-scores were utilized to assess the overall MLTC performance. The MLTC was performed three times on the annotated patient comments samples.

\begin{figure}[H]
    \centering
    \includegraphics[scale=0.35]{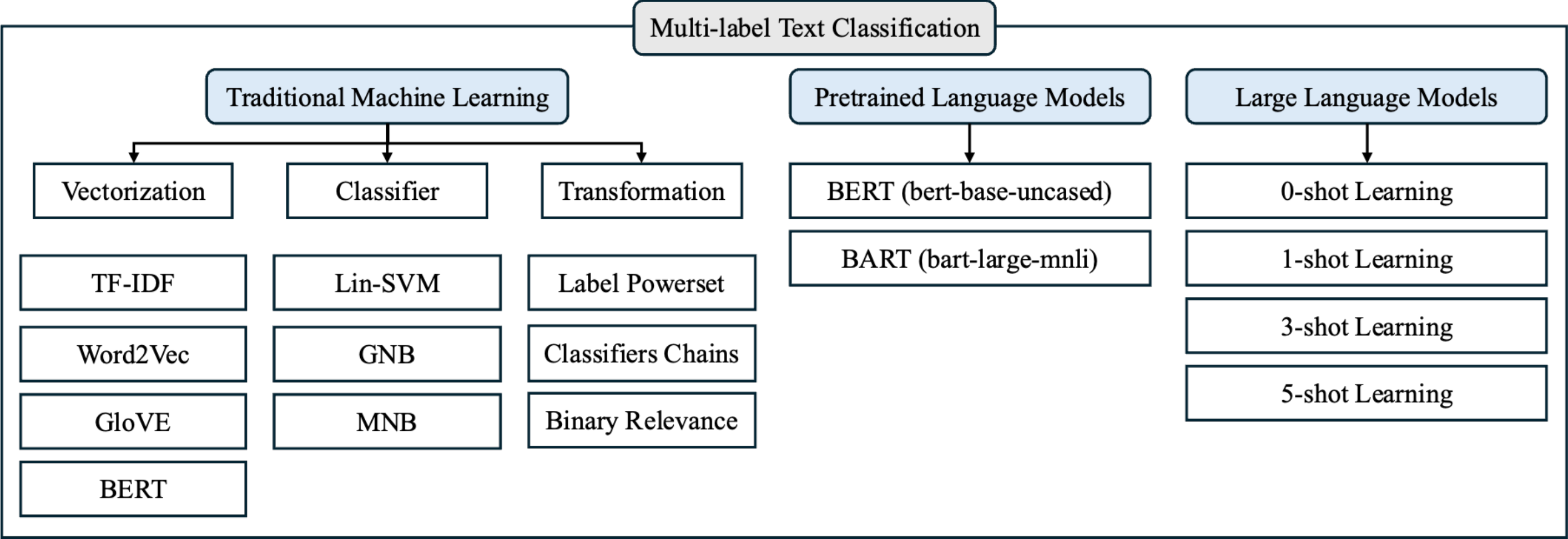}
    \caption*{\textbf{Figure 5.} MLTC Approaches}
    \label{fig:mltc-approaches}
\end{figure}

\section{Results and Discussion}
In this section, the comparison between the LLM-based, PLM-based, and traditional machine learning-based MLTC is done. Furthermore, the associations between the topics resulting from the classification of patients' comments, responses to the HCAHPS multiple-choice questions, and collected information on patients' demographics and hospital stays are statistically analyzed, emphasizing the analysis of the association between Overall Rating and Classification Topics. 
\subsection{Classification Evaluation}
First, a topic-based evaluation is carried out using F1-score and AUC-score. Tables 2 and 3 summarize the metrics values in the case of each approach.
 
\begin{table}[H]
\captionsetup{labelformat=empty}
\centering
\renewcommand{\arraystretch}{1.2}
\caption*{\textbf{Table 2.} Traditional Machine Learning and Pre-trained Language Models Results}
\begin{tabular}{c c c c c c c}
\hline
\textbf{Supervised Learning}& \multicolumn{2}{c}{\makecell{\textbf{TF-IDF + Lin-SVM} \\ \textbf{(Label Powerset)}}} & \multicolumn{2}{c}{\textbf{BERT}} & \multicolumn{2}{c}{\textbf{BART}} \\
\hline
 \textbf{Topics/Metrics} & \textbf{F1} & \textbf{AUC} & \textbf{F1} & \textbf{AUC} & \textbf{F1} & \textbf{AUC} \\
\hline
\multicolumn{1}{l}{\textbf{Positive Feedback}} & 80.00\% & 89.53\% & 33.21\% & 52.14\% & 78.84\% & 90.78\% \\
\multicolumn{1}{l}{\textbf{Noisy Environment}} & 8.33\% & 52.17\% & 7.52\% & 49.31\% & 41.71\% & 92.03\% \\
\multicolumn{1}{l}{\textbf{Missing Personal Belongings}} & 0.00\% & 50.00\% & 2.58\% & 47.78\% & 17.96\% & 93.63\% \\
\multicolumn{1}{l}{\textbf{Miscellaneous}} & 52.76\% & 71.70\% & 30.24\% & 49.40\% & 8.96\% & 51.61\% \\
\multicolumn{1}{l}{\textbf{Staff-related Issues}} & 47.20\% & 65.79\% & 28.28\% & 49.11\% & 41.75\% & 62.41\% \\
\multicolumn{1}{l}{\textbf{Long Waiting Time}} & 36.99\% & 61.57\% & 20.70\% & 49.69\% & 55.14\% & 84.82\% \\
\multicolumn{1}{l}{\textbf{Issues with Food Service}} & 55.64\% & 70.26\% & 14.41\% & 51.26\% & 72.56\% & 90.98\% \\
\multicolumn{1}{l}{\textbf{Room-related Issues}} & 61.34\% & 73.84\% & 20.67\% & 51.70\% & 57.14\% & 79.86\% \\
\multicolumn{1}{l}{\textbf{Medical-related Issues}} & 8.79\% & 52.30\% & 13.58\% & 50.13\% & 20.90\% & 65.99\% \\
\multicolumn{1}{l}{\textbf{Discharge-related Issues}} & 0.00\% & 50.00\% & 8.07\% & 48.86\% & 25.99\% & 86.60\% \\
\hline
\end{tabular}
\end{table}

The traditional machine learning model selected is TF-IDF with Lin-SVM (Label Powerset), while the PLMs chosen are BERT and BART. The selection of the conventional approach resulted from experimenting with Word2Vec, GloVe, and BERT in terms of vectorization techniques, Gradient Naïve Bayes (GNB) and MNB in terms of classifiers, and Classifiers Chains and Binary Relevance in terms of problem transformation methods. TF-IDF + Lin-SVM (Label Powerset) outperforms all these approaches using a five-fold iterative stratified cross-validation. Regarding PLMs, BERT (bert-base-uncased) and BART (bart-large-mnli) were also experimented with using a probability threshold of 0.5 for the latter. These results show that using a supervised learning method is highly sensitive to class imbalance, as demonstrated by TF-IDF + Lin-SVM (Label Powerset) performance while using small and less robust language models, even when discriminating between the classes (the case BART) relatively accurately still lack consistency across all the topics.

\begin{table}[H]
\captionsetup{labelformat=empty}
\centering
\renewcommand{\arraystretch}{1.2}
\caption*{\textbf{Table 3.} GPT-4 Turbo In-Context Learning Results}
\begin{tabular}{c c c c c c c c c}
\hline
\textbf{In-Context Learning} & \multicolumn{2}{c}{\textbf{0-shot}} & \multicolumn{2}{c}{\textbf{1-shot}} & \multicolumn{2}{c}{\textbf{2-shot}} & \multicolumn{2}{c}{\textbf{3-shot}} \\
\hline
 \textbf{Topics/Metrics} & \textbf{F1} & \textbf{AUC} & \textbf{F1} & \textbf{AUC} & \textbf{F1} & \textbf{AUC} & \textbf{F1} & \textbf{AUC} \\
\hline
\multicolumn{1}{l}{\textbf{Positive Feedback}} & \makecell{84.14\% \\ ± 0.052} & \makecell{93.40\% \\ ± 0.012} & 82.19\% & 93.26\% & \textbf{85.15\%} & 94.34\% & 83.48\% & 93.90\% \\

\multicolumn{1}{l}{\textbf{Noisy Environment}} & \makecell{\textbf{82.85\%} \\ ± 0.088} & \makecell{87.61\% \\ ± 0.055} & 77.11\% & 84.54\% & 70.89\% & 80.19\% & 76.54\% & 83.50\% \\
\multicolumn{1}{l}{\textbf{Missing Personal Belongings}} & \makecell{81.56\% \\ ± 0.133}	& \makecell{ 99.70\% \\ ± 0.002} & \textbf{88.24\%} & 99.81\% & 85.71\% & 99.77\% & 85.71\% & 99.77\% \\
\multicolumn{1}{l}{\textbf{Miscellaneous}} & \makecell{65.16\% \\ ± 0.060} & \makecell{75.18\% \\ ± 0.026} & \textbf{69.57\%} & 77.45\% & 68.67\% & 77.24\% & 68.54\% & 77.03\% \\
\multicolumn{1}{l}{\textbf{Staff-related Issues}} & \makecell{67.70\% \\ ± 0.046} & \makecell{78.25\% \\ ± 0.036} & \textbf{71.63\%} & 82.33\% & 69.28\% & 81.32\% & 70.30\% & 81.44\% \\
\multicolumn{1}{l}{\textbf{Long Waiting Time}} & \makecell{53.12\% \\ ± 0.173} & \makecell{70.80\% \\ ± 0.083} & \textbf{56.46\%} & 71.18\% & 58.37\% & 72.03\% & 57.97\% & 71.70\% \\
\multicolumn{1}{l}{\textbf{Issues with Food Service}} & \makecell{\textbf{89.01\%} \\ ± 0.027}	& \makecell{97.43\% \\ ± 0.008} & 85.00\% & 96.45\% & 85.28\% & 95.99\% & 84.42\% & 95.88\% \\
\multicolumn{1}{l}{\textbf{Room-related Issues}} & \makecell{\textbf{82.91\%} \\ ± 0.017} & \makecell{92.01\% \\ ± 0.007} & 81.52\% & 92.11\% & 81.80\% & 92.37\% & 81.32\% & 92.41\% \\
\multicolumn{1}{l}{\textbf{Medical-related Issues}} & \makecell{50.22\% \\ ± 0.034} &	\makecell{77.29\% \\ ± 0.031} & 53.15\% & 80.46\% & 49.78\% & 78.75\% & \textbf{53.64\%} & 80.89\% \\
\multicolumn{1}{l}{\textbf{Discharge-related Issues}} & \makecell{\textbf{65.57\%} \\ ± 0.019} & \makecell{95.11\% \\ ± 0.020} & 63.83\% & 93.66\% & 64.18\% & 92.72\% & 64.71\% & 93.71\% \\
\hline
\end{tabular}
\end{table}

Table 3 summarizes the performance metrics for different few-shot learning scenarios using GPT-4 Turbo. The scenarios are evaluated on the ten predefined topics using F1 and AUC scores’ means and corresponding standard deviation resulting from the three different samples annotated for the case of 0-shot learning. They include 0-shot, where the LLM performs without providing specific comments classification examples, in addition to 1-shot, 3-shot, and 5-shot scenarios, where examples are included in the prompt for better guidance. The table shows that the performance varies depending on the topic and the scenario; however, despite providing examples in the prompt that led to the performance improvement of several cases, no consistency across all topics can be deduced. The 0-shot learning performs best in cases of topics like ‘Noisy Environment’, ‘Issues with Food Service’, ‘Room-related Issues’, and ‘Discharge-related Issues’, which are more straightforward and clear topics, neither subject to further interpretations nor encompassing multiple aspects. 1-shot learning also exhibits the highest F1-scores for multiple topics, including ‘Missing Belongings’, ‘Staff-related Issues’, and ‘Long Waiting Time’. Regarding the topics, ‘Positive Feedback,’ ‘Noisy Environment,’ ‘Missing Personal Belongings,’ ‘Issues with Food Service,’ and ’Room-related Issues’ demonstrate high F1-scores exceeding 80\%. ‘Missing Personal Belongings’ shows an exceptional AUC score (>99\%). In contrast, ‘Long Waiting Time’ and ‘Medical-related Issues’ resulted in relatively lower F1-scores than other topics. It is worth mentioning that experiments with Chain-of-Thought (COT) Prompting were also conducted, but since the results were not satisfactory, they were not reported.

\begin{table}[h!]
\captionsetup{labelformat=empty}
\centering
\renewcommand{\arraystretch}{1.2}
\caption*{\textbf{Table 5.} Overall MLTC Results}
\begin{tabular}{c c c c c c}
\hline
& & \textbf{Example-based} & \multicolumn{3}{c}{\textbf{Label-based}} \\
\hline
\textbf{Category} & \textbf{Approach/Metrics} & \textbf{F1} & \textbf{Micro-F1} & \textbf{Macro-F1} & \textbf{Weighted-F1} \\
\hline
\textbf{Traditional ML} & \makecell{\textbf{TFIDF + Lin-SVM} \\ \textbf{(Label Powerset)}} & 56.80\% & 54.10\% & 35.11\% & 49.63\% \\
\hline
\multirow{2}{*}{\textbf{PLM}} & \textbf{BERT} & 17.52\% & 18.55\% & 17.93\% & 24.00\% \\
 & \textbf{BART} & 42.88\% & 43.69\% & 42.09\% & 46.00\% \\
\hline
\multirow{4}{*}{\textbf{LLM}} & \textbf{GPT-4-Turbo 1-Shot} & 74.67\% & 73.37\% & \textbf{72.87\%} & 72.86\% \\
 & \textbf{GPT-4-Turbo 3-Shot} & 74.33\% & 73.04\% & 71.91\% & 72.63\% \\
 & \textbf{GPT-4-Turbo 5-Shot} & 74.40\% & 73.35\% & 72.66\% & 72.78\% \\
& \textbf{GPT-4-Turbo 0-Shot} & \textbf{76.12\%} & \textbf{74.43\%} & 72.22\% & \textbf{73.61\%} \\
\hline
\end{tabular}
\end{table}

The results suggest that the LLM-based classifications of patient feedback are strong and significantly outperform both traditional machine learning and PLMs approaches. GPT-4 Turbo, especially in the 0-shot scenario, demonstrate that this general-purpose LLM has a strong inherent understanding of the patient comments context and is able to conduct an effective classification without neither further fine-tuning nor in-context learning especially that 0-shot learning performs interestingly well in the case of multiple topics, sometimes even outperforming few-shot learning. Furthermore, minimal variation across all four metrics can be remarked between the different few-shot scenarios which might indicate that the LLM’s base knowledge is sufficient for this MLTC and including examples in the prompt may just increase the number of input tokens without necessarily significantly ameliorating the performance. 

\subsection{Multiple-choice Questions’ Answers \& Patient Demographics vs Classification Topics}
An association analysis is conducted to understand how different classification topics expressed in patient experience feedback are related to the quantifiable HCAHPS responses and general information about the patients regarding their hospital stay and demographics. The analysis contrasts each one of the topics (i.e., a binary variable) that results from the text classification with the collected quantifiable variables (i.e., categorical variables). The quantifiable variables are categorized into three sets: patient demographics and hospital stay-related, HCHAPS multiple-choice questions answers, and HCAHPS ratings. 
\\[0.25cm]
Because we are dealing only with categorical variables, the Chi-square test of independence is used to evaluate the associations. Based on the resulting p-value and considering a significance level of 5\%, each association was judged to be statistically significant or not. While the Chi-square test allows us to conclude whether there is a statistically significant association between variables, Cramer’s V enables us to measure its strength. The interpretation of Cramer’s V values is based on Cohen (2013) and Kim (2017) and depends on the corresponding degree of freedom.  As a result, a more nuanced understanding of each association is achieved. 

\begin{landscape}

\begin{table}[h!]
\captionsetup{labelformat=empty}
\centering
\renewcommand{\arraystretch}{1.2}
\caption*{\textbf{Table 6.} MLTC Outputs and Patients Demographics \& Hospital Stay Association Analysis}
\begin{tabular}{l c c c c c c c c c c}
\toprule
 & \makecell{\textbf{Positive} \\ \textbf{Feedback}} & \makecell{\textbf{Noisy} \\ \textbf{Environment}} & \makecell{\textbf{Missing Personal} \\ \textbf{Belongings}} & \makecell{\textbf{Miscellaneous}} & \makecell{\textbf{Staff} \\ \textbf{Issues}} & \makecell{\textbf{Long Waiting} \\ \textbf{Time}} & \makecell{\textbf{Issues with} \\ \textbf{Food Service}} & \makecell{\textbf{Room} \\ \textbf{Issues}} & \makecell{\textbf{Medical} \\ \textbf{Issues}} & \makecell{\textbf{Discharge} \\ \textbf{Issues}} \\
 \hline
\textbf{Building} & \greencheck &  &  &  & \orangecheck & & & \orangecheck & & \\
\hline
\textbf{ER Admission} & & & & & & \orangecheck & & & & \orangecheck \\
\hline
\makecell[l]{\textbf{Higher grade/school} \\ \textbf{completed}} & & & \orangecheck & & & & & & & \\
\hline
\makecell[l]{\textbf{Language mainly} \\ \textbf{spoken at home}} & \orangecheck & & & & & & & & & \\
\hline
\textbf{Specialty} & \greencheck & \orangecheck & & \orangecheck & & \orangecheck & & & & \\
\hline
\textbf{Sex} & & & & & & & & & & \redcheck \\
 
\bottomrule
\end{tabular}
\end{table}

\begin{table}[h!]
\captionsetup{labelformat=empty}
\centering
\renewcommand{\arraystretch}{1.2}
\caption*{\textbf{Table 7.} MLTC Outputs and HCAHPS Ratings Association Analysis }
\begin{tabular}{l c c c c c c c c c c}
\toprule
 & \makecell{\textbf{Positive} \\ \textbf{Feedback}} & \makecell{\textbf{Noisy} \\ \textbf{Environment}} & \makecell{\textbf{Missing Personal} \\ \textbf{Belongings}} & \makecell{\textbf{Miscellaneous}} & \makecell{\textbf{Staff} \\ \textbf{Issues}} & \makecell{\textbf{Long Waiting} \\ \textbf{Time}} & \makecell{\textbf{Issues with} \\ \textbf{Food Service}} & \makecell{\textbf{Room} \\ \textbf{Issues}} & \makecell{\textbf{Medical} \\ \textbf{Issues}} & \makecell{\textbf{Discharge} \\ \textbf{Issues}} \\
 \hline
\makecell[l]{\textbf{Overall Health} \\ \textbf{Self-Rating}} & \greencheck &  &  &  & & \orangecheck & & & & \\

\hline
\makecell[l]{\textbf{Mental Health} \\ \textbf{Self-Rating}} & \orangecheck & & \orangecheck & & \orangecheck & & & & & \\

\hline
\textbf{Day Shift Rating} & \bluecheck & & \orangecheck & \orangecheck & \bluecheck & & & & \greencheck & \orangecheck \\

\hline
\textbf{Night Shift Rating} & \bluecheck & \orangecheck & \greencheck & & \bluecheck & \orangecheck & & & \greencheck & \greencheck \\

\hline
\textbf{Cleanliness Rating} & \bluecheck & \orangecheck & & & \greencheck & & & \bluecheck & \greencheck & \\

\hline
\textbf{Quietness Rating} & \bluecheck & \greencheck & & &\greencheck & \orangecheck & & \greencheck & & \\

\hline
\textbf{Overall Rating} & \bluecheck & & \greencheck & \orangecheck & \bluecheck & \greencheck & & \greencheck & \bluecheck & \orangecheck \\

\hline
\makecell[l]{\textbf{Recommendation } \\ \textbf{Rating}} & \bluecheck & & \orangecheck & & \bluecheck & \greencheck & \orangecheck & \orangecheck & \bluecheck & \orangecheck \\

\hline
\makecell[l]{\textbf{Nurses Treatment} \\ \textbf{Rating}} & \bluecheck & & \orangecheck & & \bluecheck & & & & \greencheck & \orangecheck \\

\hline
\makecell[l]{\textbf{Doctor Treatment} \\ \textbf{Rating}} & \bluecheck & & & & \greencheck & \orangecheck & \orangecheck & & \greencheck & \orangecheck \\

\hline
\textbf{Food Rating} & \bluecheck & \orangecheck & & \orangecheck & \greencheck & \orangecheck & \bluecheck & & & \\
 
\bottomrule
\end{tabular}
\end{table}

\begin{table}[h!]
\captionsetup{labelformat=empty}
\centering
\renewcommand{\arraystretch}{1.2}
\caption*{\textbf{Table 8.} MLTC Outputs and HCAHPS Multiple-Choice Questions Answers Association Analysis }
\begin{tabular}{l c c c c c c c c c c}
\toprule
 & \makecell{\textbf{Positive} \\ \textbf{Feedback}} & \makecell{\textbf{Noisy} \\ \textbf{Environment}} & \makecell{\textbf{Missing Personal} \\ \textbf{Belongings}} & \makecell{\textbf{Miscellaneous}} & \makecell{\textbf{Staff} \\ \textbf{Issues}} & \makecell{\textbf{Long Waiting} \\ \textbf{Time}} & \makecell{\textbf{Issues with} \\ \textbf{Food Service}} & \makecell{\textbf{Room} \\ \textbf{Issues}} & \makecell{\textbf{Medical} \\ \textbf{Issues}} & \makecell{\textbf{Discharge} \\ \textbf{Issues}} \\
 \hline
\makecell[l]{\textbf{Visited by} \\ \textbf{Nursing} \\  \textbf{Management}} & \greencheck &  \orangecheck &  &  & \orangecheck & \orangecheck & & & & \orangecheck \\

\hline
\makecell[l]{\textbf{Nurses Careful} \\ \textbf{Listening}} & \bluecheck & & \orangecheck & \orangecheck & \bluecheck & \orangecheck & & & \greencheck & \\

\hline
\makecell[l]{\textbf{Nurses Using} \\ \textbf{Layman Terms}} & \bluecheck & & \orangecheck & \orangecheck & \bluecheck & \orangecheck & & & \greencheck & \orangecheck  \\

\hline
\textbf{Responsiveness} & \bluecheck & \orangecheck & & \orangecheck & \bluecheck & \orangecheck & & \orangecheck & \greencheck & \\
\hline

\makecell[l]{\textbf{Toileting} \\ \textbf{Assistance}} & \greencheck & & & & \greencheck & & & & \greencheck & \\

\hline
\makecell[l]{\textbf{Doctors Using} \\ \textbf{Layman Terms}} & \bluecheck & & & \orangecheck & \greencheck & \orangecheck & & & \greencheck & \greencheck \\

\hline
\makecell[l]{\textbf{Medicine Utility} \\ \textbf{Explanation}} & \greencheck & & & & \greencheck & & & & \orangecheck & \\

\hline
\makecell[l]{\textbf{Medicine Side} \\ \textbf{Effect Description}} & \bluecheck & & & & \greencheck & & & & \greencheck & \\

\hline
\makecell[l]{\textbf{Future Needed} \\ \textbf{Help Discussion}} &  \orangecheck & \orangecheck & & & \orangecheck & \orangecheck & & & \orangecheck & \orangecheck \\

\hline
\makecell[l]{\textbf{Symptoms to} \\ \textbf{Look for} \\ \textbf{Received}} & \orangecheck & & & & \orangecheck & & & & \orangecheck & \orangecheck \\

\hline
\makecell[l]{\textbf{Preferences} \\ \textbf{Considered}} & \bluecheck  & & \orangecheck & & \greencheck & \orangecheck & & & \greencheck & \greencheck \\

\hline
\makecell[l]{\textbf{Understanding} \\ \textbf{of Health} \\ \textbf{Managing}} & \bluecheck & & & & \bluecheck & \orangecheck & & & \greencheck & \greencheck \\

\hline
\makecell[l]{\textbf{Understood } \\ \textbf{Taking Meds} \\ \textbf{Purpose}} & \bluecheck & & \orangecheck & & \greencheck & \greencheck & & & \greencheck & \orangecheck \\

\hline
\makecell[l]{\textbf{Staff Hands} \\ \textbf{Washing} \\ \textbf{Before Exam}} & \bluecheck & \orangecheck & \orangecheck & \greencheck & \greencheck & \greencheck & & \orangecheck & \greencheck & \orangecheck \\
\hline
\end{tabular}
\end{table}

\begin{table}[h!]
\begin{center}
\begin{tabular}{cccc}
\redcheck & \orangecheck & \greencheck & \bluecheck \\
Negligible to Small & Small to Medium & Medium to Large & Very Large
\end{tabular}
\end{center}
\end{table}

\end{landscape}

Tables 6, 7, and 8 summarize and visualize the different associations in addition to their strengths. If a tick is included when mapping the topic to the other variables, this indicates that the association was found to be statistically significant. Otherwise, it implies that the Chi-square test found no significant association. These color-coded ticks represent the degree of association. Red means that the association, despite being significant, is negligible to small, while orange means small to medium, green refers to medium to large, and finally, the tick is blue when it’s very large. 
\\[0.25cm]
Very strong associations were found between two topics (i.e., ‘Positive Feedback’ and ‘Staff-related Issues’) and the HCHAPS multiple-choice questions answers and ratings, respectively. The topic with the largest number of significant associations is ‘Positive Feedback’ while the one with the least number of significant associations is ‘Issues with Food Service’. The only variables that the topic ‘Positive Feedback’ doesn’t have any association with are ‘Emergency Room Admission’ (‘ER Admission’ [Y/N]), ‘Highest grade or school completed’ (categorical), and ‘Sex’. Closely following this positive topic when it comes to the associations count, we have ‘Long Waiting Time’ and ‘Medical-related Issues’, while the topic ‘Issues with Food Service’ only has statistically significant associations with three variables. Two were relatively anticipated: ‘Recommendation Rating’ (small to medium) and ‘Food Rating’ (very large). However, the third, ‘Doctor Treatment Rating’ (small to medium) was not expected. Only one association is characterized by being negligible to small: the one that contrasts ‘Discharge-related Issues’ and ‘Sex’. It can be concluded that positive feedback comments are more likely to be significantly associated with other aspects of the reported patient experience.

\subsection{Overall Rating vs Classification Topics}

\begin{table}[h!]
\captionsetup{labelformat=empty}
\centering
\renewcommand{\arraystretch}{1.2}
\caption*{\textbf{Table 9.} Regression \& Correlation Analyses Summary }
\begin{tabular}{ccccc}
\hline
\multirow{2}{*}{\makecell[c]{\textbf{Topics}}} & \multicolumn{2}{c}{\textbf{Ordinal Logistic Regression Summary}} &  \multicolumn{2}{c}{\textbf{Point-biserial}}  \\
\cline{2-5}
& \textbf{Coefficient} & \textbf{p-value} & \textbf{Correlation} & \textbf{p-value} \\
\hline
\makecell[l]{\textbf{Positive Feedback}} & 0.9686 & 0.000 & 0.418 & 0.000 \\
\makecell[l]{\textbf{Noisy Environment}} & -0.5091 & 0.006 & -0.085 & 0.005 \\
\makecell[l]{\textbf{Missing Personal Belongings}} & -0.4602 & 0.047 & -0.100 & 0.001 \\
\makecell[l]{\textbf{Miscellaneous}} & -0.2345 & 0.036 & -0.045 & 0.143 \\
\makecell[l]{\textbf{Staff-related Issues}} & -0.4114 & 0.000 & -0.287 & 0.000 \\
\makecell[l]{\textbf{Long Waiting Time}} & -0.2752 & 0.002 & -0.158 & 0.000 \\
\makecell[l]{\textbf{Issues with Food Service}} & -0.1011 & 0.370 & -0.002 & 0.950 \\
\makecell[l]{\textbf{Room-related Issues}} & -0.3610 & 0.000 & -0.191 & 0.000 \\
\makecell[l]{\textbf{Medical-related Issues}} & -0.5148 & 0.000 & -0.266 & 0.000 \\
\makecell[l]{\textbf{Discharge-related Issues}} & -0.2065 & 0.082 & -0.122 & 0.000 \\
\hline
\end{tabular}
\end{table}

The ‘Overall Rating’ variable is critical, because it forms the basis for calculating the Top Box metric, which is crucial in healthcare. This is usually done by converting the overall rating into a binary variable, where ratings of 9 or 10 are considered 1 and all other ratings 0 (Mikaeili et al., 2020). An analysis is conducted to understand the relationship between the ‘Overall Rating’ and each classification topic represented by binary variables. Because the ’Overall Rating’ is an ordinal categorical variable, Point-biserial is used for correlation analysis, while Ordinal Logistic Regression is used to model the relationship between the dependent variable (’Overall Rating’) and the independent variables (the classification topics). The outputs of this analysis are as expected. Because the only topic with a positive connotation is ‘Positive Feedback’, it is the only one with a positive coefficient and correlation. In contrast, the other topics are all characterized by negative values because they all reflect patients’ complaints. Based on the Ordinal Logistic Regression p-values, considering a significance level of 5\%, all topics are statistically significant except ‘Issues with Food Service’ and ‘Discharge-related Issues’ and their respective coefficients, which are smaller than the others’, indicate no substantial impact on the ‘Overall Rating’. Regarding Point-biserial correlation, the topics with non-significant correlation are ‘Miscellaneous’ and ‘Issues with Food Service’. Moreover, despite being statistically significant, the correlations between the ’Overall Rating’ and the other topics are weak, except for ‘Positive Feedback’.
\\[0.25cm]
This analysis reveals significant associations between patient satisfaction, which are reflected by the ‘Overall Rating’, and various aspects of the hospital stay represented by the classification topics. On the one hand, ‘Positive Feedback’ strongly correlates with better healthcare service outcomes, which suggests that positive experiences significantly impact satisfaction and high ratings. On the other hand, topics such as ‘Noisy Environment’, ‘Missing Personal Belongings’, ‘Miscellaneous’, ‘Staff-related Issues’, ‘Long Waiting Time’, ‘Room-related Issues’, ‘Medical-related Issues’, and ‘Discharge-related Issues’, show a significant negative impact and therefore indicate the areas in need of improvement. However, ‘Issues with Food Service’ demonstrates negligible effect on patients’ ‘Overall Rating’ and can be given less priority regarding resource allocation dedicated to process improvement. These insights provide a reliable foundation for healthcare professionals to enhance patients' experiences by listening to their voices.

\section{Conclusion and Future Directions}
This research paper demonstrates that leveraging LLMs, such as GPT-4 Turbo, shows significant promise in conducting patient feedback MLTC. The comprehensive analysis comparing traditional machine learning, PLMs, and in-context learning LLMs scenarios reveals the outperformance of LLMs across all the evaluation metrics and the consistency across the three considered samples demonstrated by the 0-shot scenario standard deviations. Moreover, implementing a Protected Health Information (PHI) Detection framework before feeding the patient experience feedback to the LLMs ensures that the sensitive nature of the data is handled securely and the patient’s privacy is protected. The analysis that contrasts topics’ classification, patient demographics, and responses to the multiple-choice questions reveals significant associations among various variables. This allows for the extraction of meaningful insights and enhances the understanding of the patient experience, facilitating more informed actions to improve their journey at the hospital. These findings showcase important implications for the healthcare domain since LLMs can enable automatic and enhanced patient feedback understanding and analysis. However, variability across topics’ performance suggests the need for continued research and refinement. Future work can include focusing on challenging topics and perhaps more granular topics classification and investigating LLM-based to build sophisticated classification systems. 

\section*{Acknowledgement}
The authors thank Bahareh Mahdavisharif, Yi-En Tseng, and Thuy Duong Pedicone for their diligent contribution in annotating the data samples. Their careful attention to detail and commitment to accuracy were instrumental in ensuring the quality of the data labeling.

\bibliographystyle{unsrt}  
\bibliography{references}  
[1]	Doyle, C., Lennox, L., \& Bell, D. (2013). A systematic review of evidence on the links between patient experience and clinical safety and effectiveness. BMJ Open, 3(1).
\newline
[2] 	Boulding, W., Glickman, S. W., Manary, M. P., Schulman, K. A., \& Staelin, R. (2011). Relationship between patient satisfaction with inpatient care and hospital readmission within 30 days. The American Journal of Managed Care, 17(1), 41-48.
\newline
[3] 	Senot, C., Chandrasekaran, A., \& Ward, P. T. (2016). The impact of combining conformance and experiential quality on hospitals' readmissions and cost performance. Management Science, 62(3), 829-848.
\newline
[4] 	Meterko, M., Wright, S., Lin, H., Lowy, E., \& Cleary, P. D. (2010). Mortality among patients with acute myocardial infarction: the influences of patient-centered care and evidence-based medicine. Health Services Research, 45(5p1), 1188-1204.
\newline
[5] 	Shenoy, A. (2021). Patient safety from the perspective of quality management frameworks: a review. Patient Safety in Surgery, 15(1), 12.
\newline
[6] 	Centers for Medicare \& Medicaid Services (CMS). (2018). Hospital Value-Based Purchasing. Retrieved from CMS website.
\newline
[7] 	Greaves, F., Ramirez-Cano, D., Millett, C., Darzi, A., \& Donaldson, L. (2013). Use of sentiment analysis for capturing patient experience from free-text comments posted online. Journal of Medical Internet Research, 15(11), e239.
\newline
[8] 	Manary, M. P., Boulding, W., Staelin, R., \& Glickman, S. W. (2013). The patient experience and health outcomes. The New England Journal of Medicine, 368(3), 201-203.
\newline
[9] 	Gallan, A. S., Niraj, R., \& Singh, A. (2022). Beyond HCAHPS: Analysis of patients’ comments provides an expanded view of their hospital experiences. Patient Experience Journal, 9(1), 159-168.
\newline
[10] 	 Shovkun, M. (2018). Manual and automatic analysis of patients values and preferences using Seton HCAHPS surveys (Doctoral dissertation).
\newline
[11] 	 Huppertz, J. W., \& Otto, P. (2018). Predicting HCAHPS scores from hospitals’ social media pages: A sentiment analysis. Health care management review, 43(4), 359-367.
\newline
[12] 	 Huppertz, J. W., \& Smith, R. (2014). The value of patients' handwritten comments on HCAHPS surveys. Journal of Healthcare Management, 59(1), 31-47.
\newline
[13] 	 Paul, R., Pandit, A., \& Bhardwaj, R. (2022). Transforming Healthcare through Sentiment Analysis: Tool for Patient Satisfaction. Journal of Algebraic Statistics, 13(3), 3962-3980.
\newline
[14] 	OpenAI. (OpenAI, 2024a). Enterprise Privacy. Retrieved June 8, 2024, from https://openai.com/enterprise-privacy/
[15] 	Mera, K., \& Ichimura, T. (2008). User's Comment Classifying Method Using Self Organizing Feature Map on Healthcare System for Diabetic. In 4th International Workshop on Computational Intelligence \& Applications Proceedings: IWCIA 2008 (pp. 31-36). IEEE SMC Hiroshima Chapter.
\newline
[16] 	Alemi, F., Torii, M., Clementz, L., \& Aron, D. C. (2012). Feasibility of real-time satisfaction surveys through automated analysis of patients' unstructured comments and sentiments. Quality Management in Healthcare, 21(1), 9-19.
\newline
[17] 	Wagland, R., Recio-Saucedo, A., Simon, M., Bracher, M., Hunt, K., Foster, C., ... \& Corner, J. (2016). Development and testing of a text-mining approach to analyse patients’ comments on their experiences of colorectal cancer care. BMJ quality \& safety, 25(8), 604-614.
\newline
[18] 	Pan, Q., Li, H., Chen, D., \& Sun, K. (2018, June). Sentiment analysis of medical comments based on character vector convolutional neural networks. In 2018 IEEE Symposium on Computers and Communications (ISCC) (pp. 1-4). IEEE.
\newline
[19] 	Shah, A. M., Yan, X., Shah, S. J., \& Khan, S. (2018). Use of sentiment mining and online NMF for topic modeling through the analysis of patients online unstructured comments. In Smart Health: International Conference, ICSH 2018, Wuhan, China, July 1–3, 2018, Proceedings 6 (pp. 191-203). Springer International Publishing.
\newline
[20] 	Hussain, S., Nasir, A., Aslam, K., Tariq, S., \& Ullah, M. F. (2020). Predicting mental illness using social media posts and comments. International Journal of Advanced Computer Science and Applications, 11(12).
\newline
[21] 	Shan, Y., Zhong, Z., Che, C., Jin, B., \& Wei, X. (2021, December). Aspect-level sentiment classification of chinese patient comments based on pre-trained sentiment embedding. In 2021 IEEE International Conference on Bioinformatics and Biomedicine (BIBM) (pp. 1503-1508). IEEE.
\newline
[22] 	Asghari, M., Nielsen, J., Gentili, M., Koizumi, N., \& Elmaghraby, A. (2022). Classifying comments on social media related to living kidney donation: Machine learning training and validation study. JMIR Medical Informatics, 10(11), e37884.
\newline
[23] 	Sakai, H., Mikaeili, M., Lam, S. S., \& Bosire, J. (2023). Text Classification for Patient Experience Improvement: A Neural Network Approach. In IIE Annual Conference. Proceedings (pp. 1-6). Institute of Industrial and Systems Engineers (IISE).
\newline
[24] 	Linton, A. G., Dimitrova, V., Downing, A., Wagland, R., \& Glaser, A. (2023). Weakly Supervised Text Classification on Free Text Comments in Patient-Reported Outcome Measures. arXiv preprint arXiv:2308.06199.
\newline
[25] 	Alhazzani, N. Z., Al-Turaiki, I. M., \& Alkhodair, S. A. (2023). Text Classification of Patient Experience Comments in Saudi Dialect Using Deep Learning Techniques. Applied Sciences, 13(18), 10305.
\newline
[26] 	Sakai, H., Lam, S. S., Mikaeili, M., \& Bosire, J. (2024). Patient Experience Feedback Sentiment Analysis: Combining BERT and LSTM with Genetic Algorithm Optimization. In IISE Annual Conference. Proceedings (pp. 1-6). Institute of Industrial and Systems Engineers (IISE).
\newline
[27] 	Mæhlum, P., Samuel, D., Norman, R. M., Jelin, E., Bjertnæs, Ø. A., Øvrelid, L., \& Velldal, E. (2024). It's Difficult to be Neutral--Human and LLM-based Sentiment Annotation of Patient Comments. arXiv preprint arXiv:2404.18832.
\newline
[28] 	Babbar, R., \& Schölkopf, B. (2017, February). Dismec: Distributed sparse machines for extreme multi-label classification. In Proceedings of the tenth ACM international conference on web search and data mining (pp. 721-729).
\newline
[29] 	Bhatia, K., Jain, H., Kar, P., Varma, M., \& Jain, P. (2015). Sparse local embeddings for extreme multi-label classification. Advances in neural information processing systems, 28.
\newline
[30] 	Kong, H. J. (2019). Managing unstructured big data in healthcare system. Healthcare informatics research, 25(1), 1.
\newline
[31] 	HIPAA Journal. (2024). What is considered PHI under HIPAA? HIPAA Journal. Retrieved from https://www.hipaajournal.com/considered-phi-hipaa/
\newline
[32] 	OpenAI. (OpenAI, 2024b). GPT-4 Turbo in the OpenAI API. OpenAI Help Center. Retrieved from https://help.openai.com/en/articles/8555510-gpt-4-turbo-in-the-openai-ap
\newline
[33] 	Cohen, J. (2013). Statistical power analysis for the behavioral sciences. routledge.
\newline
[34] 	Kim, H. Y. (2017). Statistical notes for clinical researchers: Chi-squared test and Fisher's exact test. Restorative dentistry \& endodontics, 42(2), 152-155.
\newline
[35] 	Mikaeili, M., Lam, S. S., \& Bosire, J. (2020). Text-Mining Framework for Hospital Patient Experience Surveys: A Case Study. In IIE Annual Conference. Proceedings (pp. 1-6). Institute of Industrial and Systems Engineers (IISE).

\section{Appendix}
\subsection{Topics Details}

\begin{table}[H]
\captionsetup{labelformat=empty}
\centering
\renewcommand{\arraystretch}{1.2}
\caption*{\textbf{Table 11.} Topics Identified and Their Detailed Descriptions}
\begin{tabular}{ll}
\hline
\makecell{\textbf{Topic}} & \textbf{Details} \\
\hline
\textbf{Positive Feedback} & \parbox[l]{12cm}{\justify Comments expressing satisfaction or compliments about any aspect of the service, staff, or facility.} \\
\hline
\textbf{Noisy Environment} & \parbox[l]{12cm}{\justify Feedback about excessive noise levels within the hospital disrupting comfort or rest.} \\
\hline
\textbf{Missing Personal Belongings} & \parbox[l]{12cm}{\justify Complaints about lost or missing personal items during the stay.} \\
\hline
\textbf{Miscellaneous} & \parbox[l]{12cm}{\justify Any feedback that does not fit into the above categories but is still valuable for quality improvement.} \\
\hline
\textbf{Staff-related Issues} & \parbox[l]{12cm}{\justify ases where the patient interaction with staff didn't meet their expectations or when communication was missing, unclear, insufficient, in a rude manner, or not helpful. Also, situations where staff did not respond timely or adequately to patient needs or inquiries because they either were busy (due to a bottleneck in the process) or they ignored the patient. Additionally, concerns or observations about staff members not maintaining proper hygiene (e.g., Staff didn't wear gloves). And feedback on staff failing to maintain patient confidentiality or privacy.} \\
\hline
\textbf{Long Waiting Time} & \parbox[l]{12cm}{\justify General complaints about long waits to receive attention from any staff member, or in the emergency room, or in hallways for rooms, procedures, or assistance, or for the discharge process, or to be assigned a bed, for hospital-provided transportation (this includes indoors and outdoors transportation), or to be admitted into the hospital, or for scheduled medical procedures, or for food.} \\
\hline
\textbf{Issues with Food Service} & \parbox[l]{12cm}{\justify Negative feedback regarding the taste, quality, or presentation of food. Also, complaints about food being served too hot or too cold. Additionally, comments where food was not provided or available when expected or indicating a lack of options or variety in the food offered. And situations where patients needed help ordering food but did not receive adequate assistance.} \\
\hline
\textbf{Room-related Issues} & \parbox[l]{12cm}{\justify Feedback related to the physical state of the room being inadequate. Also, remarks on the cleanliness of the room/hospital being unsatisfactory. Additionally, problems or conflicts arising from sharing a room with another patient. And feedback related to the room being too hot, too cold, or having inconsistent temperatures. Including also, issues related to the hospital bed, such as comfort or functionality. In addition to, situations where patients were informed that no rooms were available upon arrival or admission.} \\
\hline
\textbf{Medical-related Issues} & \parbox[l]{12cm}{\justify Feedback mentioning any complications or unexpected issues during medical procedures (e.g., reaction). Additionally, comments on inadequate pain relief or management. Also, feedback regarding incorrect diagnosis received during the stay. And observations or concerns regarding staff members lacking necessary professional medical skills.} \\
\hline
\textbf{Discharge-related Issues} & \parbox[l]{12cm}{\justify Concerns about delays or excessive waiting for the discharge process. Also, issues related to errors or confusion over discharge paperwork or any bad discharge experience.} \\
\hline
\end{tabular}
\end{table}

\end{document}